\documentclass[letterpaper, 10 pt, conference]{ieeeconf}  

\IEEEoverridecommandlockouts                              

\usepackage{hyperref}
\usepackage{graphics} 
\usepackage{checkend}	
\usepackage{graphicx}	
\usepackage{amsmath,hyperref}
\usepackage[noabbrev]{cleveref}      
\usepackage[caption=false]{subfig}
\usepackage{epsfig} 
\usepackage{times} 
\usepackage{textcomp,gensymb}
\usepackage{mathtools}
\usepackage{makecell}
\usepackage{bm}
\usepackage[ruled]{algorithm2e}
\usepackage{textcomp}
\usepackage{multirow}
\usepackage{amssymb}
\usepackage[symbol]{footmisc}

\DeclareMathOperator*{\argmin}{arg\,min}

\usepackage{xcolor}  

\newcommand{\cp}{$\mathcal{CP}$~}
\newcommand{\n}{$\mathcal{N}$~}
\newcommand{\ttwo}{$\mathcal{T}_2$~}
\newcommand{\tone}{$\mathcal{T}_1$~}
\newcommand{\etal}{\textit{et al.}~}

\title{\LARGE \bf
Anatomical Mesh-Based Virtual Fixtures for Surgical Robots*
}

\author{Zhaoshuo Li$^{1}$, Alex Gordon$^{2}$, Thomas Looi$^{2}$, James Drake$^{2}$, Christopher Forrest$^{2}$, Russell H. Taylor$^{1}$
\thanks{*This work was supported in part by a research contract from Galen Robotics, in part by Johns Hopkins University internal funds and in part by Division of Plastic Surgery and Neurosurgery at Hospital for Sick Children. \textbf{Disclosures}: 
Russel H. Taylor is a paid consultant to Galen Robotics and has an equity interest in that company. These arrangements have been reviewed and approved by JHU in accordance with its conflict of interest policy. }
\thanks{$^{1}$Authors with Laboratory for Computational Sensing and Robotics, Johns Hopkins University, Baltimore, Maryland, USA}%
\thanks{$^{2}$Authors with the Center for Image Guided Innovation and Therapeutic Intervention lab at Hospital for Sick Children, Toronto, Ontario, Canada}%
}

\begin{document}

\maketitle
\thispagestyle{empty}
\pagestyle{empty}

\begin{abstract}

This paper presents a dynamic constraint formulation to provide protective virtual fixtures of 3D anatomical structures from polygon mesh representations. The proposed approach can anisotropically limit the tool motion of surgical robots without any assumption of the local anatomical shape close to the tool. Using a bounded search strategy and Principle Directed tree, the proposed system can run efficiently at 180 Hz for a mesh object containing 989,376 triangles and 493,460 vertices. The proposed algorithm has been validated in both simulation and skull cutting experiments. The skull cutting experiment setup uses a novel piezoelectric bone cutting tool designed for the da Vinci research kit. The result shows that the virtual fixture assisted teleoperation has statistically significant improvements in the cutting path accuracy and penetration depth control. The code has been made publicly available at \href{https://github.com/mli0603/PolygonMeshVirtualFixture}{https://github.com/mli0603/PolygonMeshVirtualFixture}.

\end{abstract}

\section{INTRODUCTION}

Craniosynostosis occurs when one or more cranial sutures prematurely fuse.  The fused cranial sutures do not allow for proper expansion of the skull as the child grows. It is estimated that the birth incidence of craniosynostosis is 0.4 per 1,000 births \cite{hunter1976craniosynostosis}. To avoid serious cognitive and cosmetic complications, these cranial sutures must be reopened by surgical intervention. This operation typically lasts for a long time (around 8 hours) and requires delicate tool movement to avoid incision-related complications \cite{tatum2008advances}.

Surgical robots, such as the da Vinci Surgical System (Intuitive Surgical, US), can mitigate the challenges by extending human capabilities. Virtual fixture (VF), i.e. software motion constraints, can further reduce the operational difficulties by allowing the surgeon and robot to work together to complete the surgical task with improved stability, reliability and precision \cite{bowyer2013active}. VF can stop the surgeon from making sudden motions and accidentally damaging critical anatomies. Examples of VF in robot-assisted surgeries where high precision is required include neurosurgery \cite{ueda2017toward} and intraocular surgery \cite{li2020hybrid}. To the best of our knowledge, the use of the da Vinci system for craniofacial surgery has never been attempted. 

Prior work has constructed complex virtual fixtures from simpler primitives in parametric forms. For example, Prada \etal  \cite{prada2005study} use a set of point VF to assist path following tasks. Part \etal \cite{park2001virtual} defines a hyper-plane aligned with an artery to assist dissection. Marinho \etal \cite{marinho2019dynamic} select points on the patient's anatomy to construct VF and the corresponding vector-field inequalities are created to ensure that the surgical tool stays sufficiently far from the anatomy. The process of selecting and defining primitives can be extremely tedious and impractical if done manually before the surgery, due to the geometric complexity of anatomical structures. 

Ren \etal \cite{ren2008dynamic} propose a VF system for cardiac surgery to protect the exterior of the patient's heart without manually defining any primitives. The proposed system first extracts a mesh representation of the heart using the Marching Cubes algorithm on the MR/CT images and then approximates the surface of the heart using a set of B-spline basis functions. While this approach works for modeling the exterior shape of the patient’s heart, the accuracy of the B-spline approximation can degrade severely for other anatomical structures as the complexity grows. Li \etal \cite{li2007spatial} overcome the previous limitation by generating VF from an anatomical mesh directly at 30 Hz in the robot control loop. One limitation of this approach, however, is the assumption that the anatomical surfaces near the tool at any given time can always produce a set of convex linear constraints. Although this assumption is often valid, it breaks down in cases where the surgical tool must work in proximity to highly complex anatomical structures.

In this paper, we propose a new dynamic constraint formulation for the online generation of virtual fixtures from polygon mesh representations of complex anatomical structures in 3D. The proposed algorithm does not make any assumptions of the local concavity/convexity of the model relative to the tool. These models can be obtained pre-operatively such as from CT \cite{santosh2019medical}, or intra-operatively such as from 3D ultrasound \cite{furuhata2014interactive}. The proposed formulation can be used as a high-level control algorithm to limit the tool's motion within a ``safe'' region. It can be also used to provide haptic feedback in a teleoperated control setting.

The contributions of the paper can be summarized as follows:
\begin{itemize}
    \item a dynamic constraint formulation for virtual fixtures of anatomies modeled by polygon meshes, without the assumption of convexity/concavity of the model in proximity to the tool,
    \item validation experiment of the proposed approach in simulation with various shapes,
    \item run time experiment of the proposed approach,
    \item VF assisted skull cutting experiments using a novel piezoelectric cutter designed for craniotomy.
\end{itemize}

The rest of this paper is organized as follows: \autoref{sec:vf_bg} provides the necessary background of VF and \autoref{sec:pm_vf} presents the proposed approach. \autoref{sec:sim_experiment} presents the validation experiment in simulation and the run time experiment. \autoref{sec:cutting_experiment} presents the skull cutting experiments. \autoref{sec:conclusion} concludes the paper.

\section{BACKGROUND}
\label{sec:vf_bg}

\subsection{Constraint Optimization}
Constraint optimization approaches are well-established methods to implement virtual fixtures. In this work, we formulate robot kinematic motion control as a quadratic optimization problem with linear constraints \cite{funda1996constrained}. The objective function solves for the desired motion as

\begin{equation} \label{eq1}
    \begin{split}
    \argmin_{\Delta \pmb{q}} & \| \Delta\pmb{x} - \Delta \pmb{x_d} \|_2, \\ 
     \text{subject to }& \pmb{A} \Delta \pmb{x} \geq \pmb{b} \\
     & \Delta \pmb{x} = \pmb{J} \Delta \pmb{q}
    \end{split}
\end{equation}
where $\Delta \pmb{x}$ and $\Delta \pmb{x_d}$ are the computed and desired incremental Cartesian positions, and $\Delta\pmb{q}$ is the incremental joint position. $\pmb{J}$ is the Jacobian matrix relating the joint space to the Cartesian space. $\pmb{A}$ and $\pmb{b}$ are matrix and vectors necessary to describe the linear constraints.

In a telemanipulated control environment, $\Delta \pmb{x_d}$ in the objective function is computed from the motion of the master manipulator. Additional objectives may be added to express additional desired behaviors. 

\begin{figure}[h]
    \centering
    \includegraphics[width=0.5\linewidth]{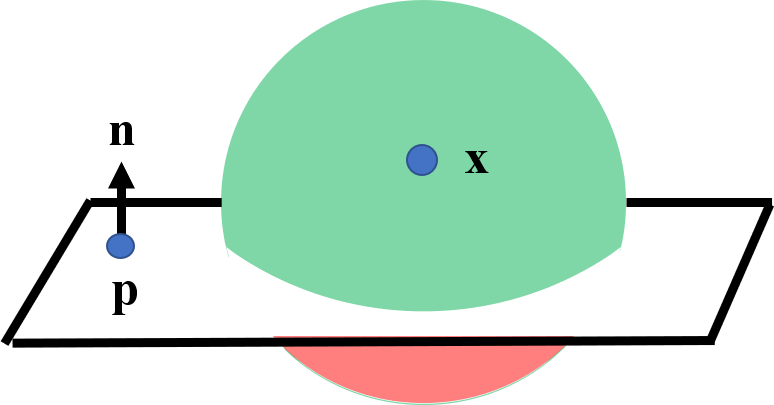}
    \caption{Illustration of plane constraint. The radius of the sphere is the maximum motion capable of the robot in one control iteration. Green zones denote the allowable region and red zones denote the forbidden region.}
    \label{fig:plane_constraint}
\end{figure}

Inequality constraints can be used to impose motion constraints. For example, a virtual forbidden wall for tool tip $\pmb{x}$ can be defined by a hyper-plane with normal $\pmb{n}$ and point $\pmb{p}$, i.e. tool tip can only move on the positive side of the hyper-plane as shown in \autoref{fig:plane_constraint}. This can be achieved by forcing the signed distance $d_t$ from the tool tip to plane at any time $t$ to be positive, i.e.,
\begin{equation}
    \begin{split}
        d_{t-1} &= \pmb{n}^T (\pmb{x}-\pmb{p}) \\
        \Delta d &= \pmb{n}^T \Delta \pmb{x} \\
        d_t &= d_{t-1} + \Delta d \geq 0 \\
        \pmb{n}^T \Delta \pmb{x} &\geq - \pmb{n}^T (\pmb{x}-\pmb{p})
        \label{eqn:plane_constraint}
    \end{split}
\end{equation}
where $\Delta d$ is the change of the signed distance. The constrained motion can be realized by setting $\pmb{A}=\pmb{n}^T$ and $\pmb{b}=- \pmb{n}^T (\pmb{x}-\pmb{p})$. The constrained least-squares problem may then be solved to produce the desired motion. Additional terms may be added to further constrain the tool motion.

\subsection{Polygon Mesh}
\begin{figure}[htpb]
    \centering
    \subfloat[]{\includegraphics[trim={0 -0.5cm 0 0.5cm},clip,width=0.24\linewidth]{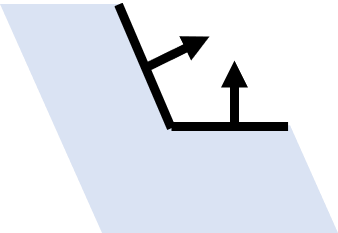}}
    \subfloat[]{\includegraphics[width=0.25\linewidth]{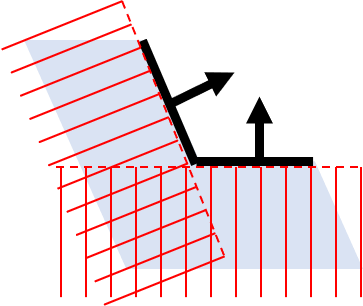}}
    \subfloat[]{\includegraphics[trim={0 -0.5cm 0 0},clip,width=0.18\linewidth]{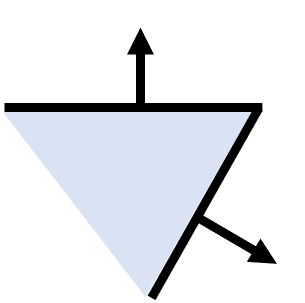}}
    \subfloat[]{\includegraphics[trim={0 0 0.6cm 0.5cm},clip,width=0.3\linewidth]{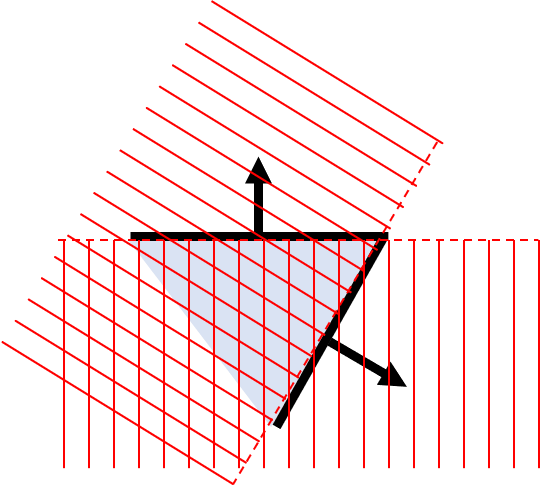}}
    
    \caption{2D illustration of (a-b) a locally \textbf{concave surface}, and (c-d) a locally \textbf{convex surface}. Red crosshatched zones denote forbidden region. Note in (d) many allowable regions are ruled out. Arrows denote the face normals. Light blue zones denote the patient's anatomy.}
    \label{fig:convex_concave}
\end{figure}

In this work, polygon meshes consisting of triangles are used to represent anatomical surfaces. A locally \textbf{concave surface} (\autoref{fig:convex_concave}a) produces a \textbf{convex set of linear constraints}. It is safe to include all triangles as plane constraints (\autoref{fig:convex_concave}b). A locally \textbf{convex surface} (\autoref{fig:convex_concave}c) produces a \textbf{non-convex set of linear constraints}. Naively adding all triangles as plane constraints will rule out many allowable regions (\autoref{fig:convex_concave}d). This necessitates an approach to dynamically activate and deactivate the constraints based on the local convexity and concavity of the anatomical surfaces.

\color{black}
\section{VIRTUAL FIXTURE FOR POLYGON MESH}
\label{sec:pm_vf}
\subsection{Polygon Mesh Constraint Algorithm}
Instead of considering the whole mesh object, a \textit{motion volume} is built around the current position defined by the maximum motion capable of the robot in one control iteration. Currently, we implement this as a motion sphere with radius corresponding to the maximum distance the robot can move in one time step. The triangles intersected by the motion sphere are the only necessary ones to be considered in the current iteration. To enable efficient geometric-search intersection, the anatomical mesh is stored as a Principle Direction Tree (PD-Tree) \cite{williams1997augmented}. The PD-Tree is similar to the KD-Tree, but with nodes split along the maximum distributive direction of the data. This provides an additional boost in search efficiency especially when triangles are not uniformly distributed. During the PD-Tree construction, the adjacency information of the triangles is stored. Using the PD-Tree of the anatomy and the motion sphere of the tool position, the corresponding closest points $\mathcal{CP}$ and face normals $\mathcal{N}$ of each intersected triangle are returned.

\begin{figure}[h]
    \centering
    \subfloat[]{\includegraphics[width=0.38\linewidth]{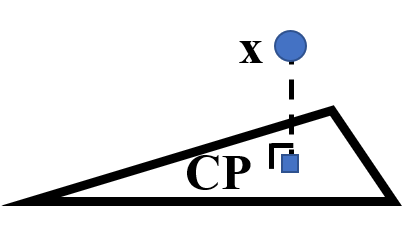}}
    \subfloat[]{\includegraphics[trim={0 0.23cm 0 0},clip,width=0.52\linewidth]{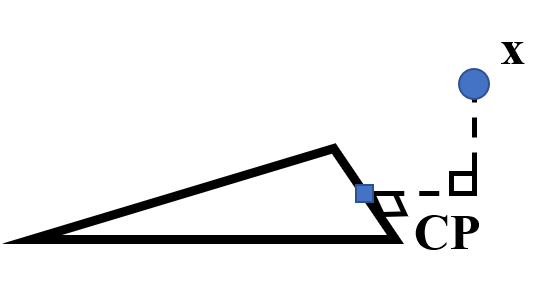}}
    \caption{$\mathcal{CP}$ location, (a) \textit{in-triangle}, (b) \textit{on-edge}. The blue rectangle denotes the $\mathcal{CP}$. The right-angle sign denotes perpendicularity.}
    \label{fig:cp_location}
\end{figure}

The closest point location on a triangle is found using \cite{jones19953d}, which can land in two places - \textit{in-triangle} or \textit{on-edge} (\autoref{fig:cp_location}). The closest points on each triangle are necessary to determine the local convexity/concavity and approximate local geometries using plane constraints. Based on the \cp location, there are only three cases for a given triangle $\mathcal{T}_i$ to be added to the list of active constraints $\mathcal{L}$. The algorithm is shown in \autoref{alg:poly_mesh_constraint}.

\begin{algorithm}[h]
    \caption{Polygon Mesh Constraint}
    \label{alg:poly_mesh_constraint}
    \SetAlgoLined
    \SetKwInOut{Input}{Input}
    \Input{~PD-Tree, Current Position~$x$}
    \KwResult{List of Active Plane Constraints $\mathcal{L}$}
    
    Find intersected triangles $\mathcal{T}$, corresponding closest points \cp and face normals \n\;
    
    \For{triangle $\mathcal{T}_i \in\mathcal{T}$}{
        \uIf{$\mathcal{CP}_i$ \textit{in-triangle} \& $\mathcal{N}_i^T (x-\mathcal{CP}_i) \geq 0$}{
            add \{$\mathcal{N}_i$, $\mathcal{CP}_i$\} to $\mathcal{L}$ \;
        }
        \uElseIf{$\mathcal{CP}_i$ \textit{on-edge}}{
            Find adjacent triangle(s) $\mathcal{T}_{i,a}$ \;
            \uIf{$\mathcal{CP}_i == \mathcal{CP}_{i,a}$ \& \textcolor{black}{locally convex}}{
                add \{$x-\mathcal{CP}_i$, $\mathcal{CP}_i$\} to $\mathcal{L}$ \;
            }
            \uElseIf{$\mathcal{N}_i^T (x-\mathcal{CP}_i) \geq 0$ \& \textcolor{black}{locally concave}}{
                add \{$\mathcal{N}_i$, $\mathcal{CP}_i$\} to $\mathcal{L}$ \;
            }
        }
     }
\end{algorithm}

\textbf{Condition 1} ($\mathcal{CP}_i$ \textit{in-triangle} \& $\mathcal{N}_i^T (x-\mathcal{CP}_i) \geq 0$): If the closest point $\mathcal{CP}_i$ is \textit{in-triangle} and the current position $x$ is on the positive side of the face normal ($\mathcal{N}_i^T (x-\mathcal{CP}_i) \geq 0$), then the triangle is added as a constraint. The face normal $\mathcal{N}_i$ and closest point $\mathcal{CP}_i$ will be used in \autoref{eqn:plane_constraint}.

On the other hand, if the closest point is \textit{on-edge}, then the information of the adjacent triangle(s) $\mathcal{T}_{i,a}$ is retrieved, where the subscript $a$ denotes ``adjacent''. When \textbf{Condition 2} or \textbf{Condition 3} is met, an active plane constraint is added. The local convexity is found by
\begin{equation}
    \begin{split}
        convexity = \begin{cases}
                        1, & \text{if } \mathcal{N}_{i,a}^T v > 0\\
                        0, & \text{if } \mathcal{N}_{i,a}^T v < 0
                    \end{cases}
    \end{split}
\end{equation}
where $v$ is the one of the non-shared edges originating from the shared vertex of the neighboring triangle.

\textbf{Condition 2} ($\mathcal{CP}_i$ \textit{on-edge} \& $\mathcal{CP}_i == \mathcal{CP}_{i,a}$ \& \textit{locally convex}): If the closest point of the adjacent triangle(s) $\mathcal{CP}_{i,a}$ is equal to the closest point of the current triangle $\mathcal{CP}_i$, then the two points are on the same shared edge. When this occurs for the locally convex surface (\autoref{fig:convex_concave}c), the edge can be approximated by a plane with the normal equal to the vector $x-\mathcal{CP}_i$, which points from the closest point $\mathcal{CP}_i$ to the current position $x$. The plane also passes the closest point $\mathcal{CP}_i$. In theory, when the $\mathcal{CP}_i$ falls on a vertex, there could be many triangles intersecting this point. However, it is sufficient to only consider any one of the two adjacent triangles who shares an edge and $\mathcal{CP}_i$ with $\mathcal{T}_{i}$. The rest of the triangles whose closet points also fall on this vertex can be safely discarded as they will result in the same constraint.

\textbf{Condition 3} ($\mathcal{CP}_i$ \textit{on-edge} \& $\mathcal{N}_i^T (x-\mathcal{CP}_i)$ \& \textit{locally concave}): If none of the above cases apply, and the current position $x$ is on the positive side of the face normal of the current triangle ($\mathcal{N}_i^T (x-\mathcal{CP}_i) \geq 0$) and the local surface is concave (\autoref{fig:convex_concave}a), the face normal $\mathcal{N}_i$ and closest point $\mathcal{CP}_i$ are used to define the plane constraint.

\subsection{Completeness of Algorithm}
In this section, we will enumerate the different local cases that can arise and explain how the algorithm handles each case. Note that the figures use a 2D representation, where line segments denote triangles, arrows denote face normals, blue squares denote closest points, and blue circles denote the current tool tip position. The motion sphere is represented by the circle containing green (allowable), red (forbidden) and gray (undecided) regions.

\begin{figure}[htpb]
    \centering
    \subfloat{\includegraphics[width=0.3\linewidth]{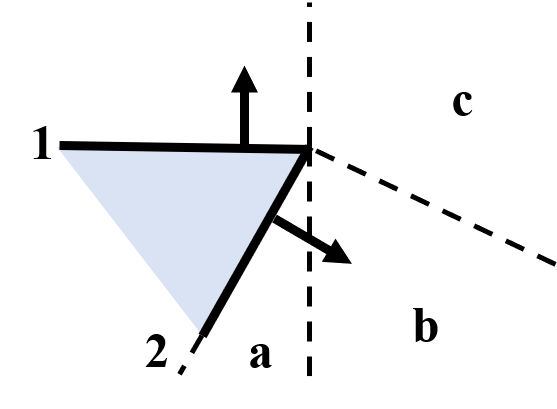}}
    \addtocounter{subfigure}{-1}
    \subfloat[]{\includegraphics[trim={0 -0.2cm 0 0},clip,width=0.27\linewidth]{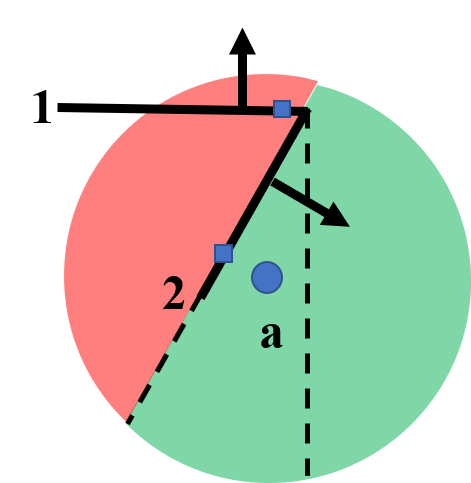}}
    \subfloat[]{\includegraphics[trim={0 -0.2cm 0 0},clip,width=0.3\linewidth]{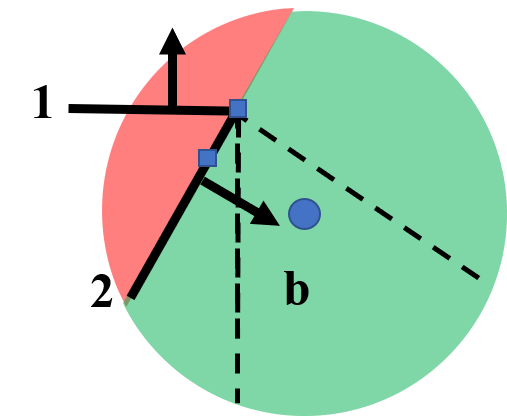}}
    \subfloat[]{\includegraphics[trim={0 0.73cm 0 0},clip,width=0.3\linewidth]{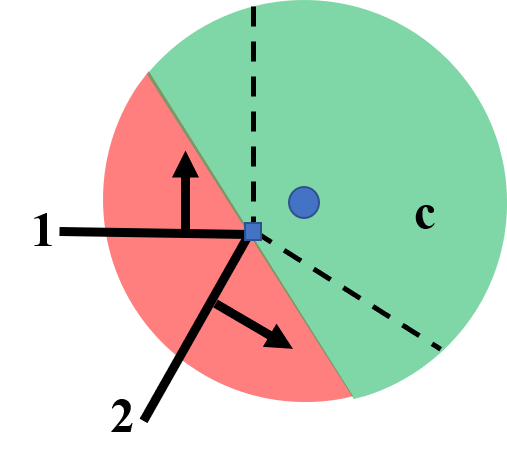}}
    \caption{Enumeration of a \textbf{locally convex} shape.}
    \label{fig:convex_enum}
\end{figure}

A locally \textbf{convex shape} formed by two triangles is illustrated in \autoref{fig:convex_enum}. There are three cases to be considered, and the rest will be the same due to symmetry. In case 4(a), \ttwo is activated (Condition 1); \tone fails to activate since \cp is \textit{in-triangle} while the current position is on the negative side of the face normal. In case 4(b), \ttwo is activated (Condition 1); \tone fails to activate since $\mathcal{CP}_1$ is \textit{on-edge} while the adjacent triangle's closest point $\mathcal{CP}_2$ is not on the same edge, nor is the current position on the positive side of both normals. In case 4(c), a plane with normal equal to the vector pointing from \cp to the current position is activated (Condition 2), since \cp of both triangles land on the same edge and form a locally convex shape.

\begin{figure}[htpb]
    \centering
    \subfloat{\includegraphics[width=0.28\linewidth]{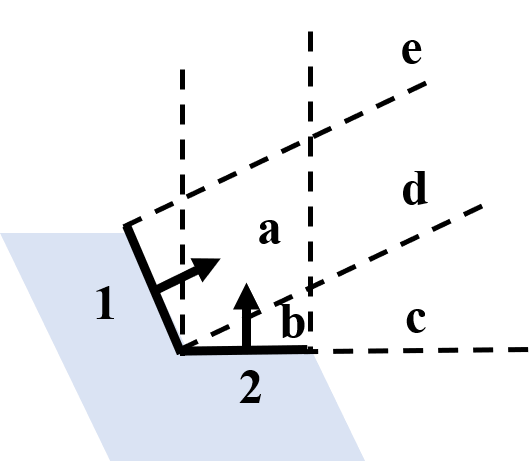}}
    \hspace{3pt}
    \addtocounter{subfigure}{-1}
    \subfloat[]{\includegraphics[width=0.25\linewidth]{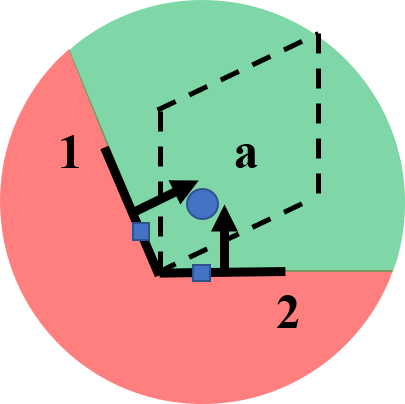}}
    \subfloat[]{\includegraphics[width=0.29\linewidth]{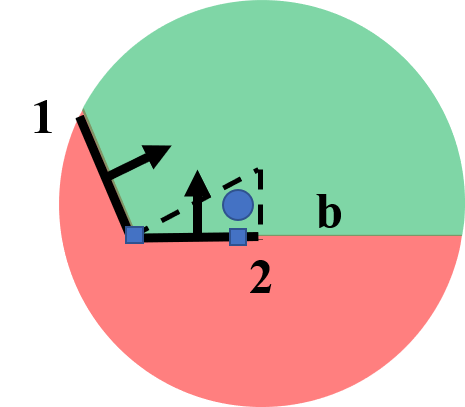}}
    
    \subfloat[]{\includegraphics[width=0.28\linewidth]{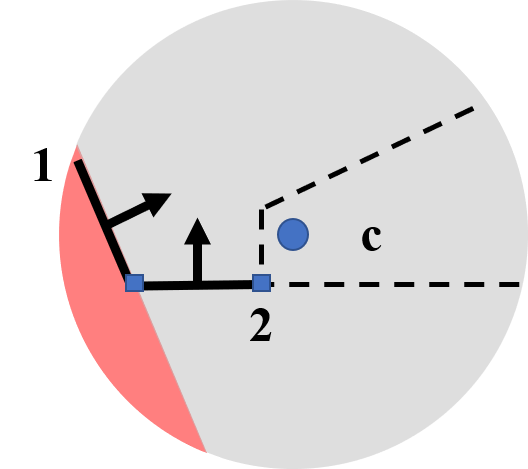}}
    \subfloat[]{\includegraphics[width=0.28\linewidth]{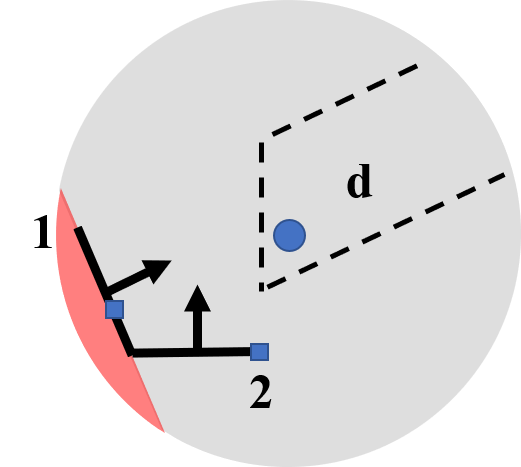}}
    \hspace{1pt}
    \subfloat[]{\includegraphics[trim={0 0.6cm 0 0},clip,width=0.255\linewidth]{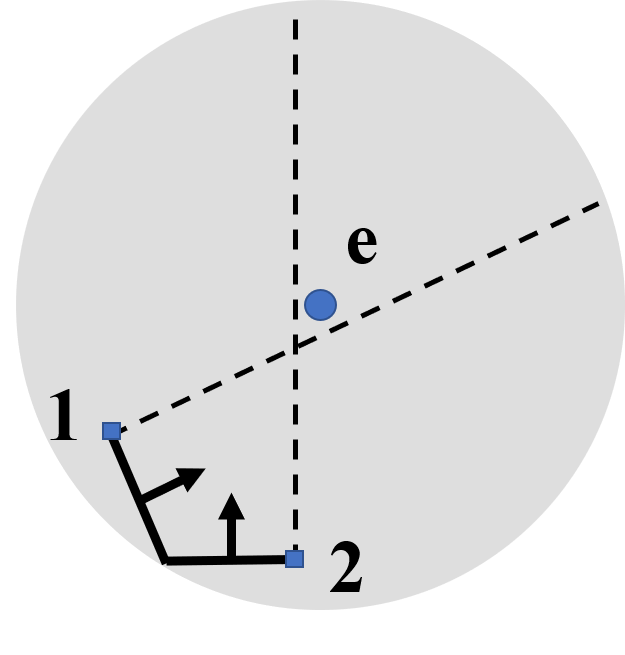}}
    \caption{Enumeration of a \textbf{locally concave} shape.}
    \label{fig:concave_enum}
\end{figure}

The case where two triangles forming a locally \textbf{concave shape} is illustrated in \autoref{fig:concave_enum}. In case 5(a), both triangles are activated (Condition 1). In case 5(b), \ttwo is activated (Condition 1) and \tone is activated due to local geometric concavity (Condition 3). In case 5(c), \tone is activated due to local geometric concavity (Condition 3); the activation of \ttwo is undecided due to the missing adjacency information of the other edge. If adjacency information is given, it will be one of the previously discussed cases. In case 5(d), \tone is activated (Condition 1); the activation of \ttwo is undecided due to the missing adjacency information. In case 5(e), both the activation of both triangles is undecided due to the missing adjacency information.

For both locally concave and convex shapes, the above cases cover all possible situations independent of the angle between the two triangles. \autoref{tab:enumerated_cases} summarizes the result.

\begin{table}[htpb]
\centering
\caption{Activated constraints of the enumerated local cases. Brackets contain the normal and point used in \autoref{eqn:plane_constraint}.}
\label{tab:enumerated_cases}
\begin{tabular}{cc||c|c|c|}
\cline{3-5}
 &  & \textbf{Condition 1} & \textbf{Condition 2} & \textbf{Condition 3} \\ \hline
\multicolumn{1}{c|}{\multirow{3}{*}{\makecell{Concave \\ \autoref{fig:concave_enum}} }} & a & \{$\mathcal{N}_2$, $\mathcal{CP}_2$\} &  &  \\ \cline{2-5} 
\multicolumn{1}{c|}{} & b & \{$\mathcal{N}_2$, $\mathcal{CP}_2$\} &  &  \\ \cline{2-5} 
\multicolumn{1}{c|}{} & c &  & \{x-$\mathcal{CP}_2$, $\mathcal{CP}_2$\} &  \\  \hline\hline
\multicolumn{1}{c|}{\multirow{5}{*}{\makecell{Convex \\ \autoref{fig:convex_enum}}}} & a & \makecell{\{$\mathcal{N}_1$, $\mathcal{CP}_1$\},\\ \{$\mathcal{N}_2$, $\mathcal{CP}_2$\}} &  &  \\ \cline{2-5} 
\multicolumn{1}{c|}{} & b & \{$\mathcal{N}_2$, $\mathcal{CP}_2$\} &  & \{$\mathcal{N}_1$, $\mathcal{CP}_1$\} \\ \cline{2-5} 
\multicolumn{1}{c|}{} & c &  &  & \{$\mathcal{N}_1$, $\mathcal{CP}_1$\} \\ \cline{2-5} 
\multicolumn{1}{c|}{} & d & \{$\mathcal{N}_1$, $\mathcal{CP}_1$\} &  &  \\ \cline{2-5} 
\multicolumn{1}{c|}{} & e &  &  &  \\ \hline
\end{tabular}
\end{table}

\section{SIMULATION EXPERIMENT}
\label{sec:sim_experiment}
\subsection{Simulation Validation}
\begin{figure}[htpb]
    \centering
    \subfloat[]{\includegraphics[trim={20cm 1cm 15cm 2cm},clip,width=0.22\linewidth]{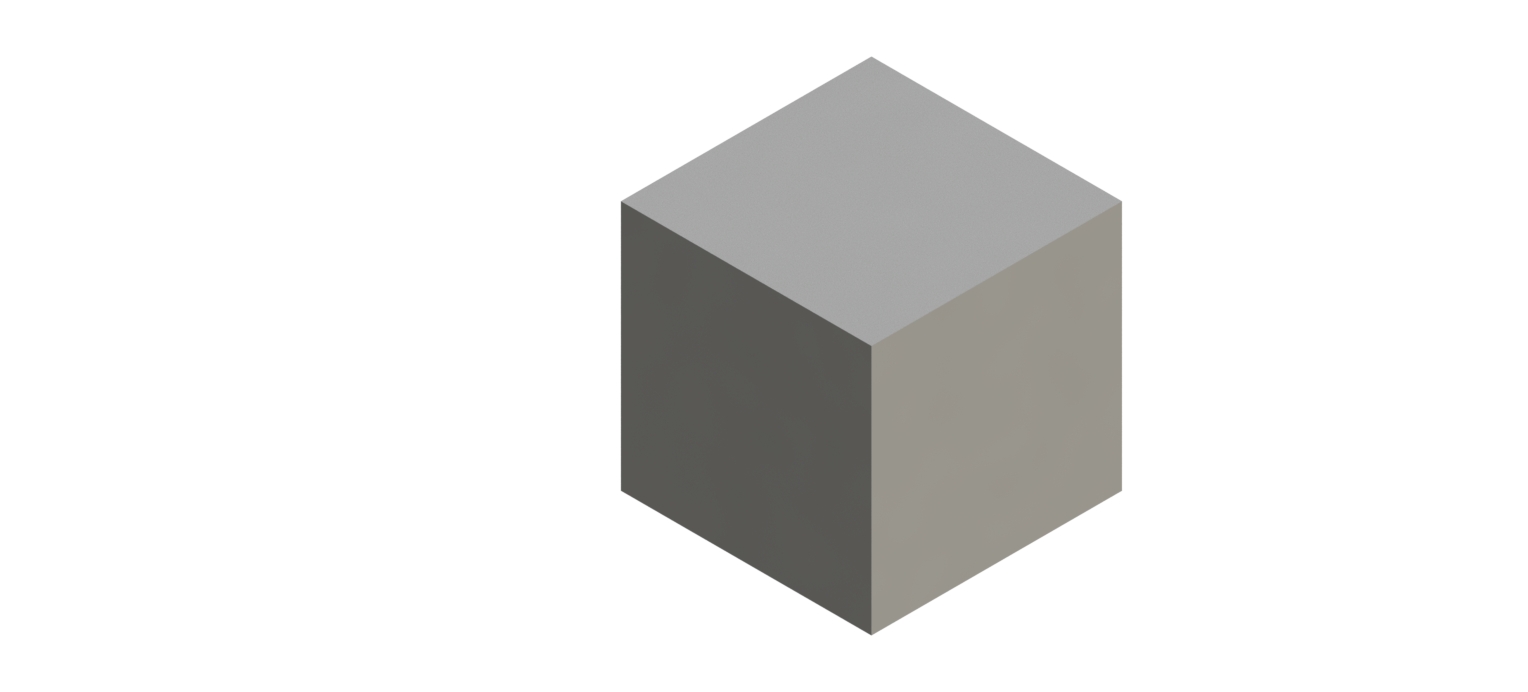}}
    \subfloat[]{\includegraphics[trim={20cm 3cm 15cm 3cm},clip,width=0.25\linewidth]{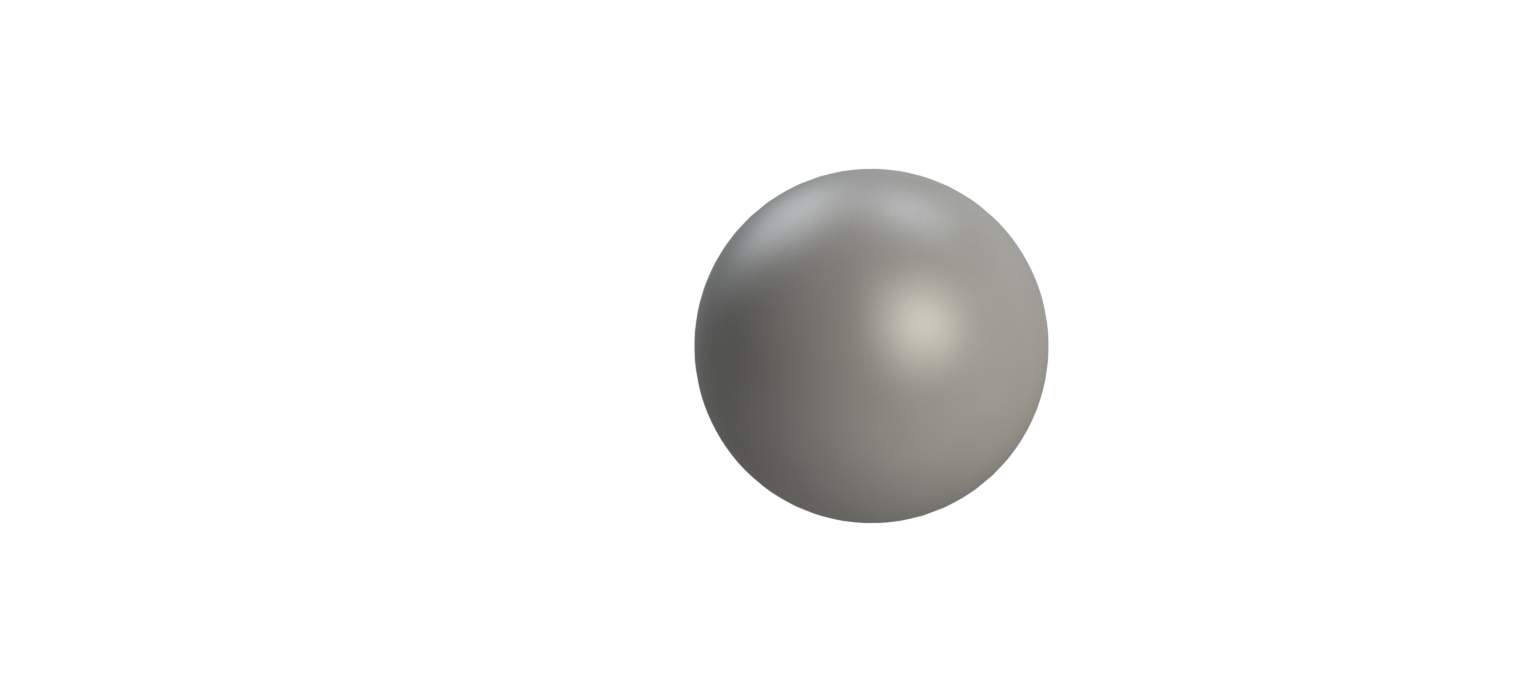}}
    
    \subfloat[]{\includegraphics[trim={18cm 5cm 15cm 2cm},clip,width=0.3\linewidth]{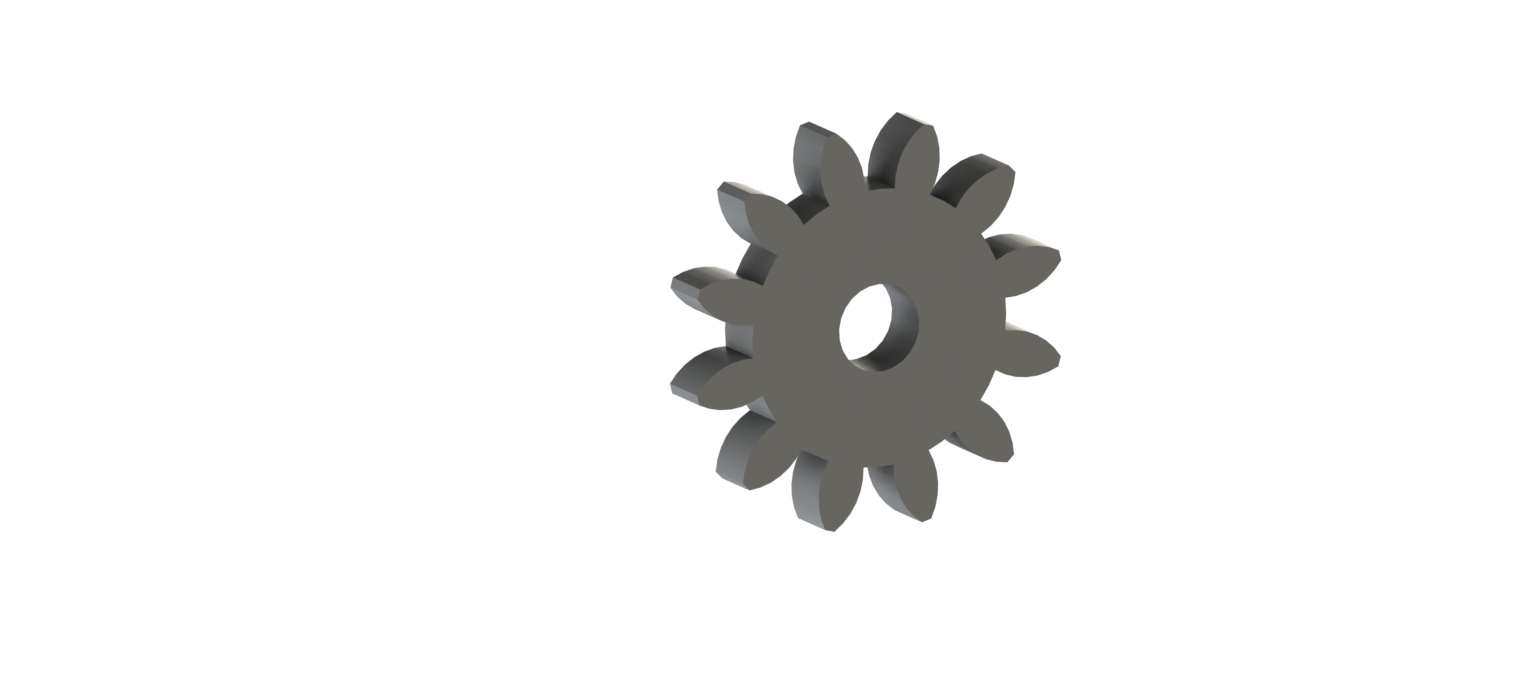}}
    \subfloat[]{\includegraphics[trim={20cm 5cm 18cm 5cm},clip,width=0.28\linewidth]{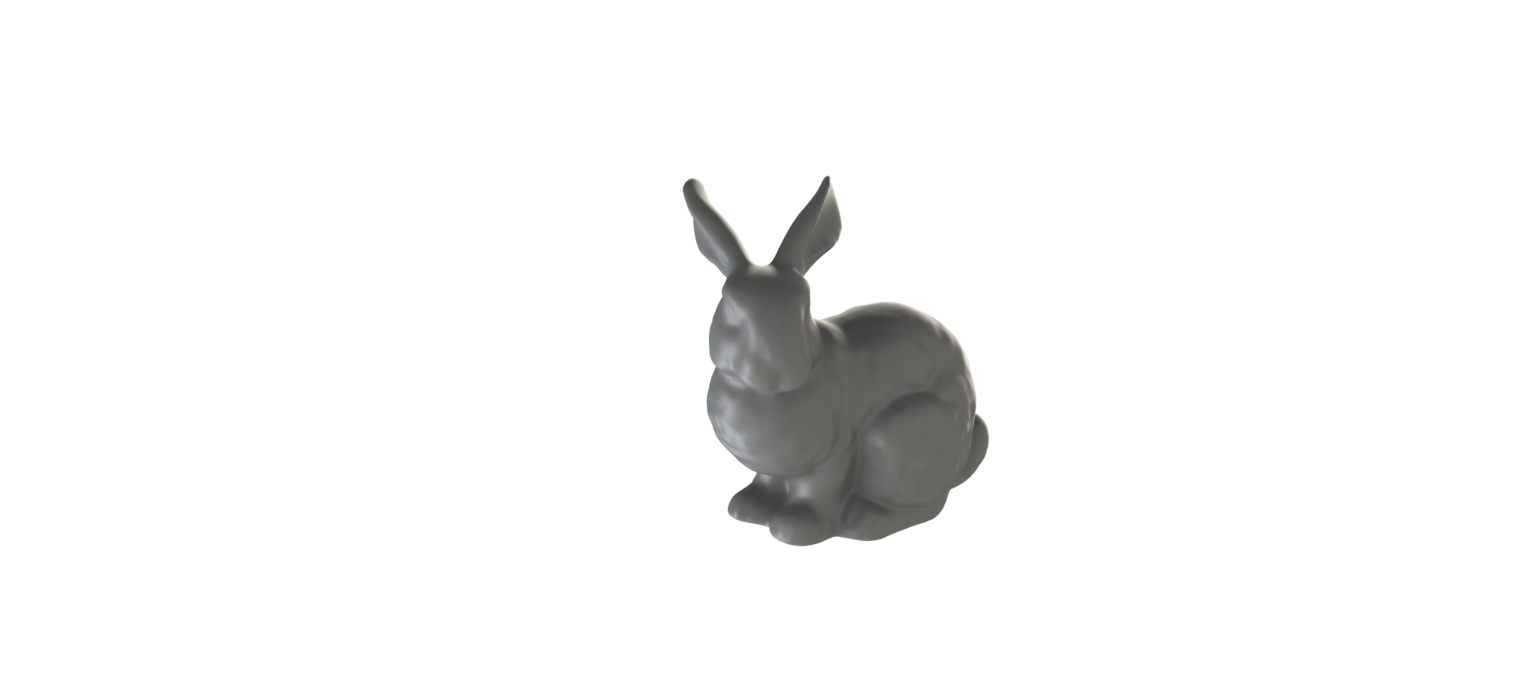}}
    \subfloat[]{\includegraphics[trim={20cm 5cm 15cm 5cm},clip,width=0.3\linewidth]{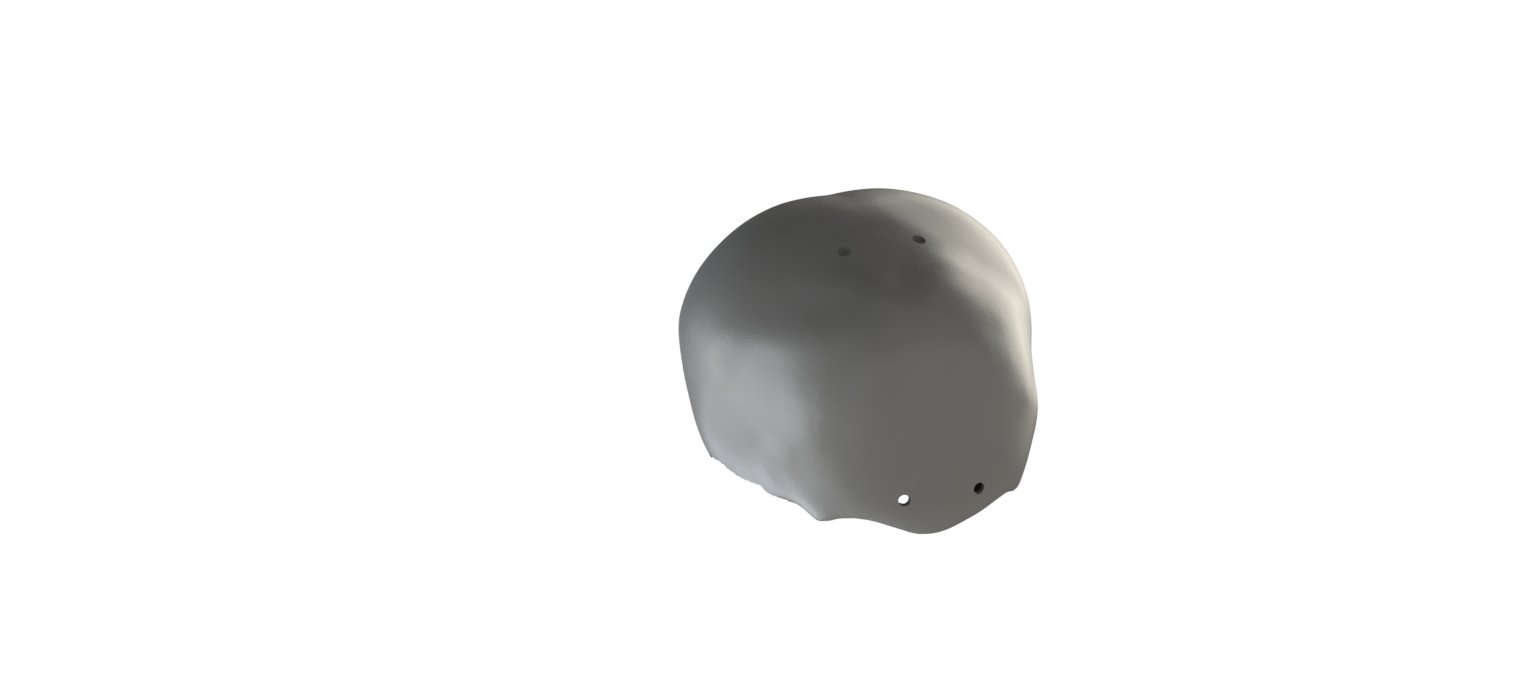}}
    \caption{Examples of the objects tested in simulation. (a) cube, (b) sphere, (c) gear, (d) Stanford bunny, (e) pediatric skull.}
    \label{fig:simulation_objects}
\end{figure}

The proposed algorithm is first evaluated in simulation. In the experiment, there are 11 shapes tested, including simple geometries, such as cube, and complex geometries such as pediatric skull. Some examples of the tested mesh objects are shown in \autoref{fig:simulation_objects}. The pediatric skull is generated from a CT scan, and later used to 3D print the skull phantoms for the cutting experiments. The proposed algorithm is successful at providing the virtual forbidden area for all the mesh objects. The full result can be found in the video supplementary material. 

\subsection{Run Time Experiment}
\begin{figure}[htpb]
    \centering
    \includegraphics[width=0.7\linewidth]{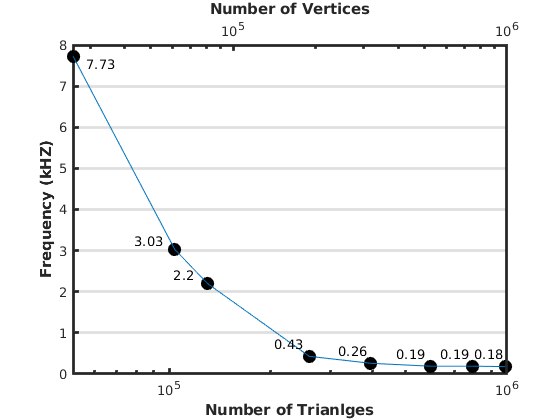}
    \caption{Run time result of Frequency (kHz) versus Number of Vertices (top, log scale) and Number of Triangles (bottom, log scale). The frequency is labeled beside each data point.}
    \label{fig:run_time_result}
\end{figure}

A run time experiment is conducted in simulation to evaluate the performance of the proposed approach. The period of the run time loop, including the time of finding intersected triangles, detecting activation of triangles and solving for optimization solution is recorded. The time is then converted to frequencies with respect to the number of triangles and vertices in the mesh objects. The algorithm is implemented in C++ using \href{https://github.com/jhu-cisst/cisst/wiki}{the CISST library} developed at Johns Hopkins University. The system used is Ubuntu 18.04 with Intel Core i7-8700. The object tested is a high-resolution Stanford bunny as shown in \autoref{fig:simulation_objects}d. As the number of triangles increases from 51,978 to 989,376 and the number of vertices increases from 25,991 to 493,460,  the run time loop frequency decreases from 7.73 kHz to 180 Hz. The performance of the algorithm exceeds the minimal requirement for a high-level teleoperated control loop of 1 kHz when the mesh object contains less than approximately 200,000 triangles. The full result is plotted in \autoref{fig:run_time_result}.

\section{SKULL CUTTING EXPERIMENT}
\label{sec:cutting_experiment}
The previously discussed VF algorithm is integrated into the da Vinci Research Kit (dVRK) \cite{kazanzides2014open}, and two experiments are conducted to validate the effectiveness of the VF assistance. The experiment setup can be found in \autoref{subsec:experiment_setup}. During the experiments, the performance using two control modes are compared - teleoperation and VF assisted teleoperation. In the path tracing experiment (\autoref{subsec:path_tracing}), a pre-cut skull phantom is traced along a planned path. The smoothness and penetration depth of the tool motion are compared. In the cutting experiment (\autoref{subsec:cutting}), four skull phantoms are cut using a novel piezoelectric cutter. The cutting accuracy, penetration depth and completion time of the tool motion are compared.

\subsection{Experiment Setup}
\label{subsec:experiment_setup}
\begin{figure}[h]
    \centering
    \subfloat[]{\includegraphics[width=0.9\linewidth]{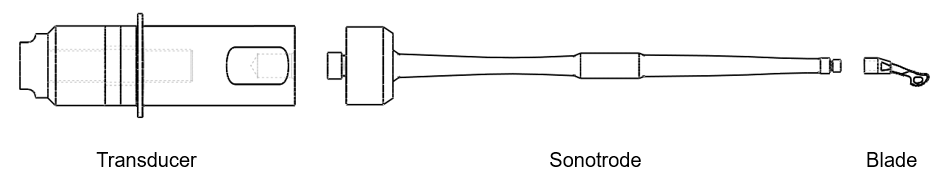}}
    
    \subfloat[]{\includegraphics[width=0.9\linewidth]{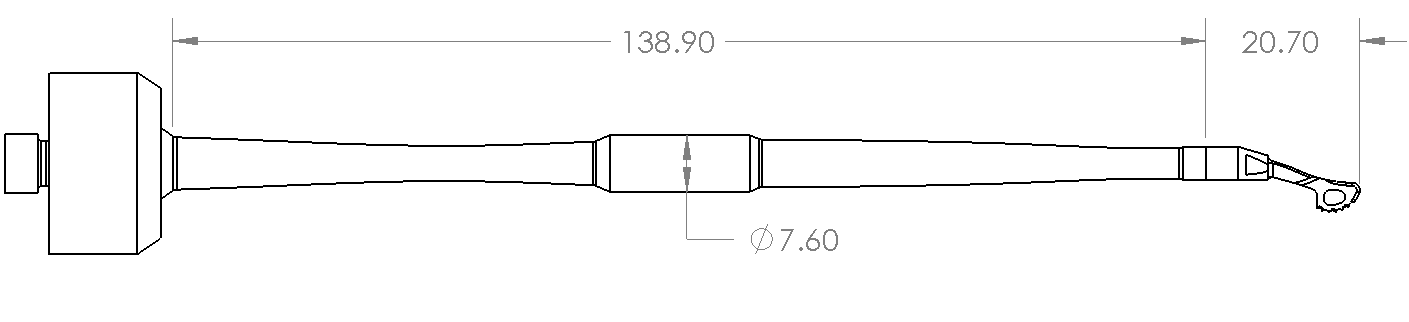}}
    
    \subfloat[]{\includegraphics[trim={0 0 0.5cm 0},clip,width=0.5\linewidth]{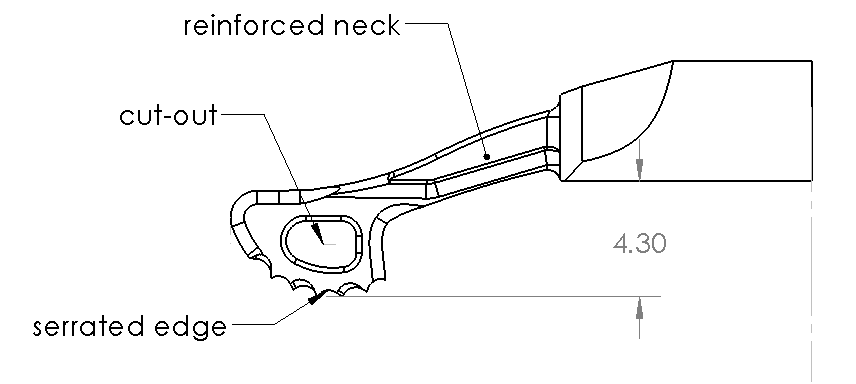}}
    \subfloat[]{\includegraphics[width=0.5\linewidth]{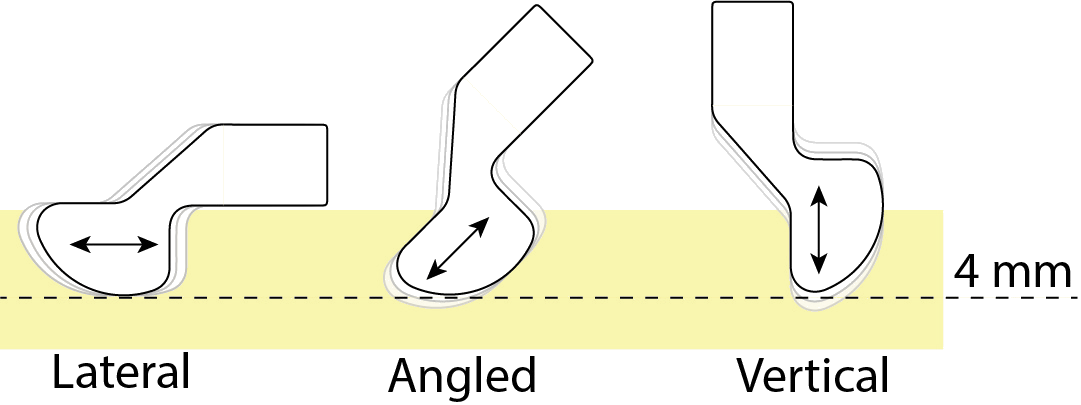}}
    
    \caption{The piezoelectric cutter used in the experiments, (a) components, (b) dimension, (c) blade design, (d) illustration of the blade cutting angles.}
    \label{fig:cutter}
\end{figure}

The piezoelectric cutter used in this experiment is an improved version from \cite{gordon2018ultrasonic}. The cutter operates at 20-30 kHz frequency, with vibration magnitude of 15-350 microns. It has three components, the transducer, the sonotrode, and the blade. The blade is designed such that it can cut along various angles for up to 4 mm depth. The design details of the piezoelectric cutter can be found in \autoref{fig:cutter}.

The skull phantoms used in the experiments are 3D printed with VisiJet PXL + ColorBond (3D Systems, US). The material has a flexural modulus of 7.16 GPa, which is similar to human skulls of 7.46 GPa \cite{motherway2009mechanical}. The skull has a uniform thickness of 2 mm. A 30 mm planned cutting path is inked on the skull on the Sagittal plane. There are six fiducial markers placed on the skull for registration.

\begin{figure}
    \centering
    \includegraphics[width=0.9\linewidth]{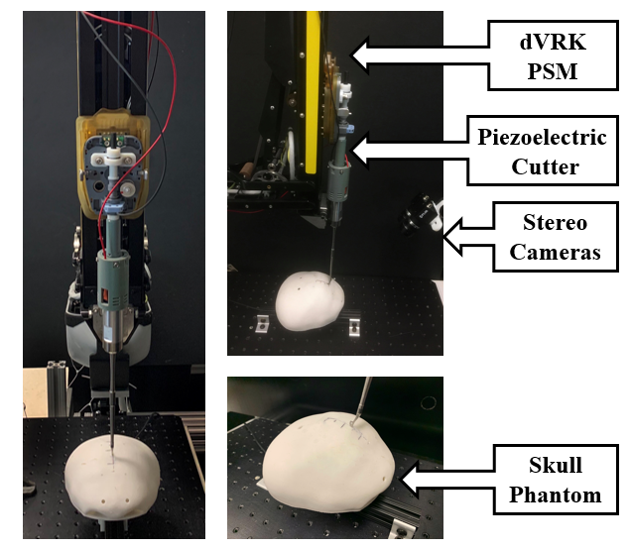}
    \caption{Experiment setup for skull cutting experiments.}
    \label{fig:experiment_setup}
\end{figure}

The cutting tool is mounted on the dVRK Patient Side Manipulator (PSM). The skull phantom is placed directly underneath the PSM, centered at the RCM point. A pair of Blackfly S cameras (FLIR Systems, Canada) is used for stereo vision.
The experiment setup is shown in \autoref{fig:experiment_setup}. 

In the VF assisted mode, haptic feedback is provided on the Master Tool Manipulator (MTM). The force exerted by the MTM on the surgeon's hand is proportional to the positional difference between the constrained PSM position and proxy position (i.e. the position without any VF constraint). The PSM is constrained such that the cutter can deviate from the planned path up to 1 mm perpendicular to the Sagittal plane on each side. The motion sphere radius for the polygon mesh constraint algorithm is set to 5 mm.

\subsection{Path Tracing}
\label{subsec:path_tracing}
\begin{figure}[htpb]
    \centering
    \subfloat[]{\includegraphics[trim={2cm 0 0 0},clip,width=0.55\linewidth]{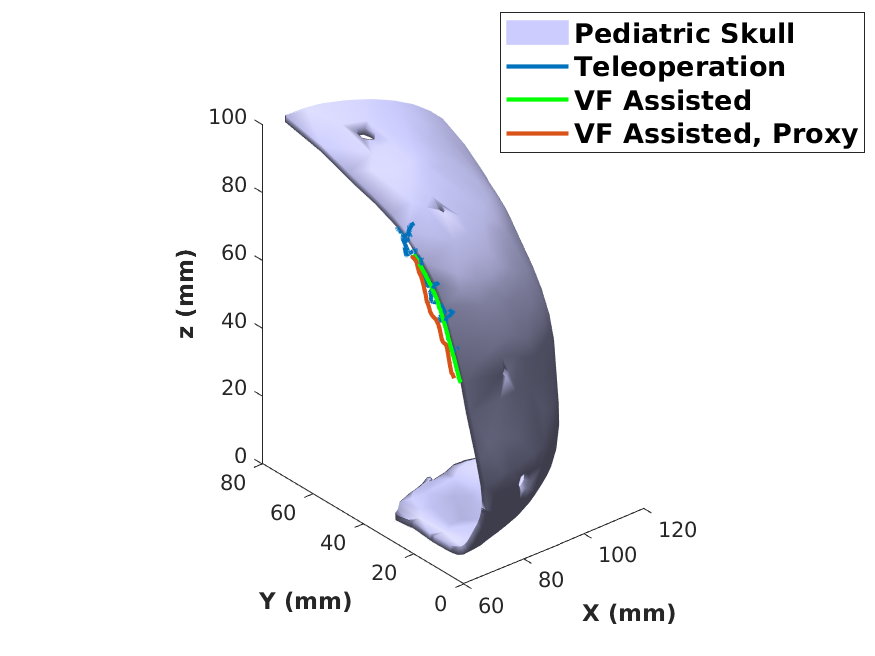}}
    
    \subfloat[]{\includegraphics[trim={2cm 0 0 0},clip,width=0.55\linewidth]{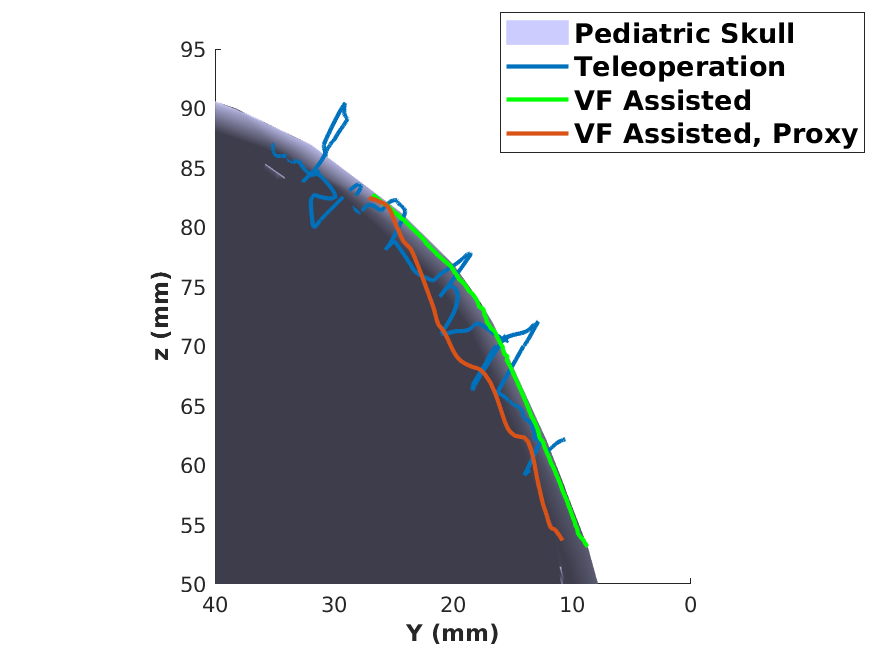}}
    \caption{Overlays of sample trajectories from the path tracing experiment on the pre-cut pediatric skull.}
    \label{fig:skull_trace}
\end{figure}

In this experiment, the exterior of a pre-cut skull is traced using the piezoelectric cutter along a planned path. The goal is to compare the penetration depth control between relying on only stereo vision and relying on stereo vision and VF assistance. The path is pre-cut to avoid excessive contact force between the cutter and the skull. The experiment is repeated five times for each mode. The overlays of the trajectories on the cross-section of the skull are shown in \autoref{fig:skull_trace}. In the VF assisted mode, the path can be traced by constantly pushing the PSM down since the VF will stop the cutter at the exterior surface of the skull. Even though the proxy trajectory has some deviation, the VF effectively constrains the motion. The trajectory of the tool without VF assistance has more variation and deviation due to the challenges in depth perception. The VF assisted trajectory is visually smoother and aligns with the exterior of the skull more accurately. 

During surgery, it is desirable to have smooth trajectories without sudden adjustments. To qualitatively evaluate the motion smoothness of the trajectories, the spectral arc length (SAL) \cite{balasubramanian2011robust} is computed:
\begin{equation}
    \begin{split}
        SAL &= -\int_{0}^{\omega_c} \sqrt{(\frac{1}{\omega_c})^2+(\frac{d\hat{V}(\omega)}{d\omega})^2}d\omega\\
        d\hat{V}(\omega) & \triangleq \frac{V(\omega)}{V(0)}
    \end{split} 
\end{equation}
where $\omega_c$ is the cut-off frequency of 20 Hz, and $V$ is the velocity of the cutter. The intuition of SAL is that in the Fourier domain, the complexity of the velocity profile in the frequency domain will be retained and amplified. Due to the negative sign in front of the integral, a larger SAL will result in a smaller arc length, thus indicating a smoother motion. The SAL scores are $-20.98\pm0.79$ and $-10.02\pm3.7$ respectively for the trajectories of teleoperation and VF assisted teleoperation modes. The improvement of the trajectory smoothness with VF assistance is statistically significant (p-value=1.9E-4).

\subsection{Cutting}
\label{subsec:cutting}
In this experiment, four skulls are cut along the planned path using the piezoelectric cutter. The interior of the skull is covered by a layer of wax to evaluate the penetration depth after cutting. Two skulls are cut with VF assistance and two without. The following metrics are used to compare the performance of teleoperation and VF assisted teleoperation:
\begin{itemize}
    \item \textbf{Cutting Path Deviation (CPD)}: the deviation of the cut path from the planned path. The minimal deviation is desired for the most accurate cutting path. 
    \item \textbf{Penetration Depth (PD)}: the penetration depth from the interior skull surface. Minimal penetration is desired to avoid damages to the dura mater.
    \item \textbf{Duration}: the total time it takes to complete the cutting procedure.
\end{itemize}

\begin{figure}[h]
    \centering
    \subfloat[]{\includegraphics[trim={0 1.5cm 0 1cm},clip,height=0.5\linewidth]{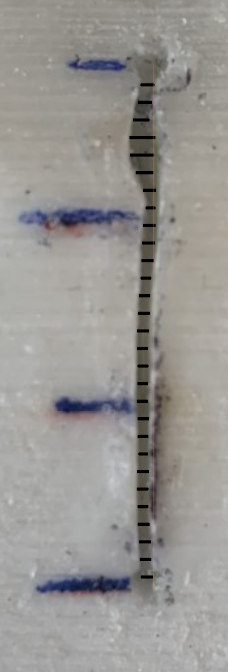}}
    \hspace{1pt}
    \subfloat[]{\includegraphics[trim={0 2.5cm 0 1cm},clip,height=0.5\linewidth]{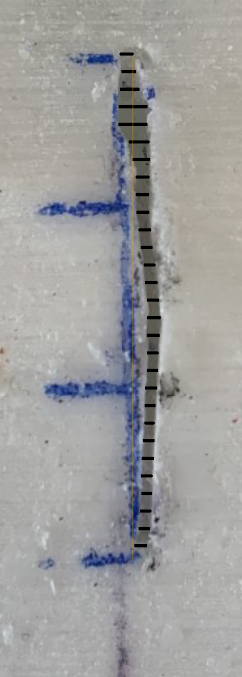}}
    \hspace{10pt}
    \subfloat[]{\includegraphics[height=0.5\linewidth]{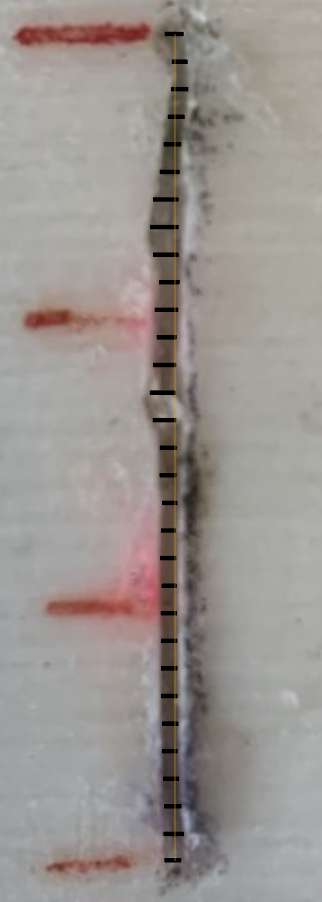}}
    \hspace{1pt}
    \subfloat[]{\includegraphics[height=0.5\linewidth]{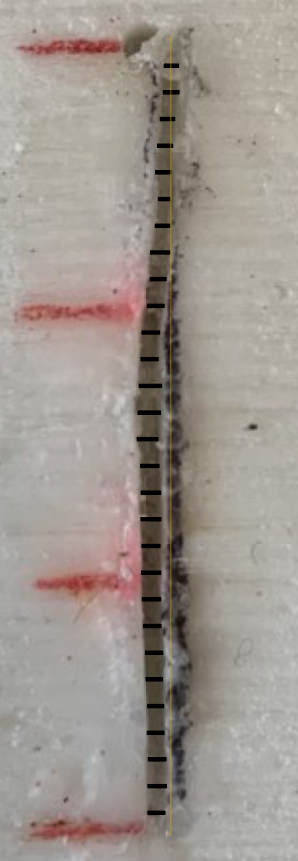}}
    \caption{Cut path using (a-b) teleoperation, (c-e) VF assisted teleoperation. Marks are inked every 10 mm along the path. The distance between the black horizontal line segments (labeled during CPD analysis) inside the path is 1 mm.}
    \label{fig:cut_path}
\end{figure}

\begin{figure}[h]
    \centering
    \subfloat[]{\includegraphics[width=0.8\linewidth]{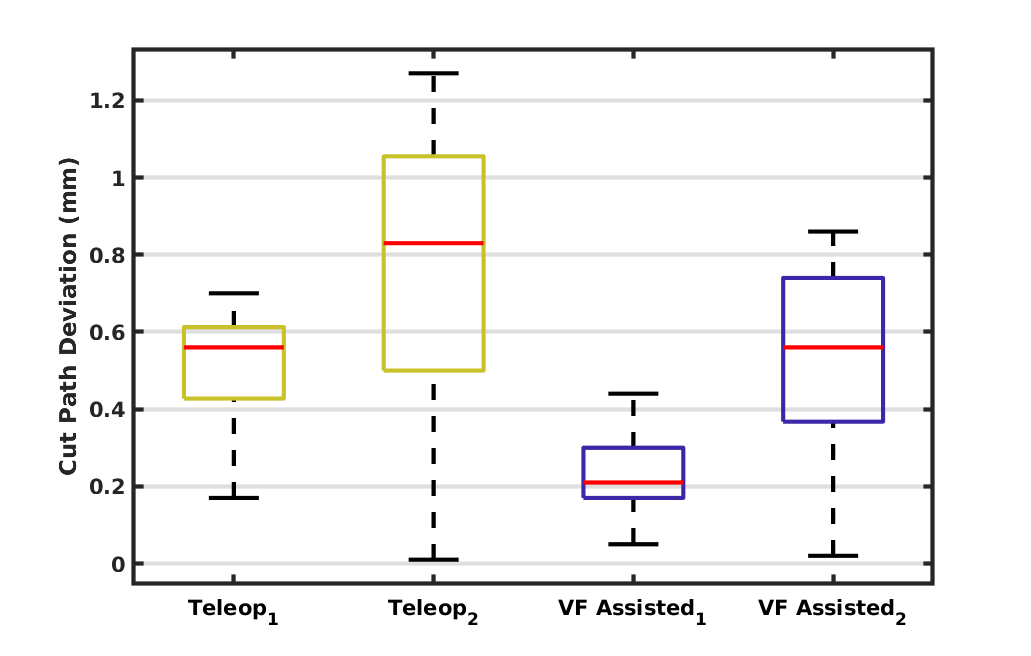}}
    
    \subfloat[]{\includegraphics[width=0.8\linewidth]{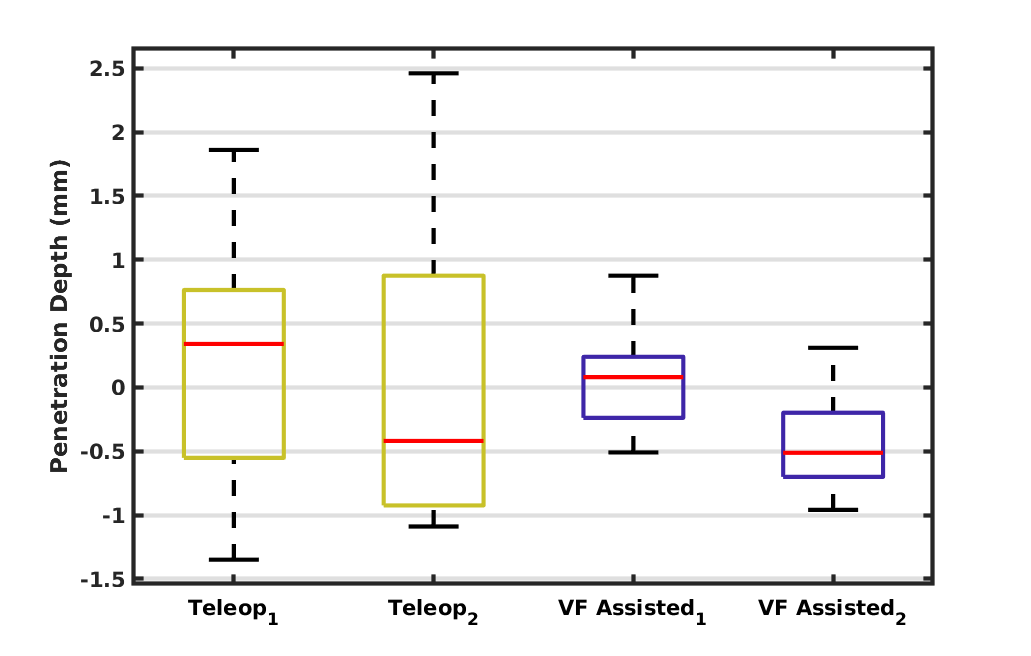}}
    \caption{Experiment results for the four trials of skull cutting. (a) Cutting path deviation. (b) Penetration depth.}
    \label{fig:experiment_result}
\end{figure}

The CPD is measured as the perpendicular distance from the furthest points of the cut path to the planned path at 1 mm intervals, totaling 30 measurements per cutting trial. To measure the CPD, the cutting path is imaged by a high-resolution camera. The pixel-to-mm conversion is obtained by dividing the pixel distance between the inked line segments by 10 mm, which is the actual distance. The CPD is measured in pixel first and converted to mm using the pixel-to-mm conversion. The result can be found in \autoref{fig:experiment_result}a and the cut paths of the four trials can be found in \autoref{fig:cut_path}. The maximum CPD of the teleoperated mode is 1.27mm, while the maximum CPD of VF assisted mode is only 0.86mm. The mean$\pm$std CPD are $0.63\pm0.30$mm for the teleoperation mode and $0.37\pm0.24$mm for the VF assisted mode. The improvement in cutting path accuracy with VF assistance has been found statistically significant. (p-value=2.3E-6).

\begin{figure}[h]
    \centering
    \includegraphics[width=0.8\linewidth]{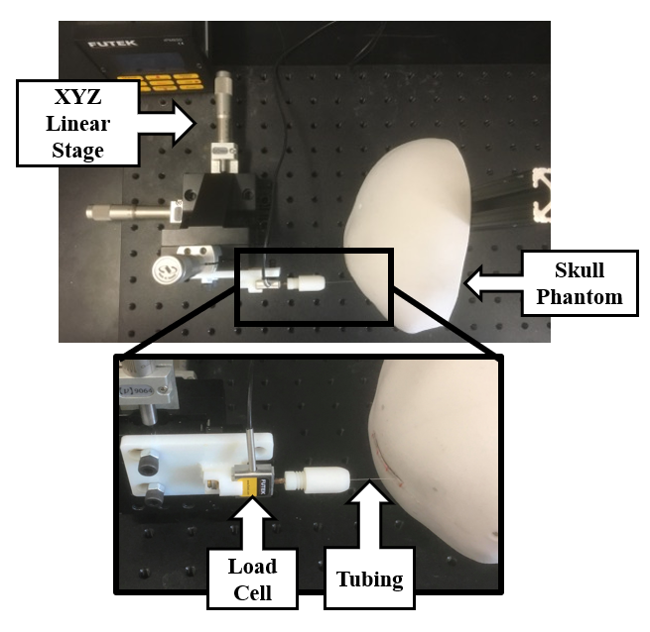}
    \caption{Penetration depth measurement setup.}
    \label{fig:pd_setup}
\end{figure}

The PD is measured as the distance between the exterior surface of the skull and the top surface of the wax minus the 2 mm skull thickness at 1 mm intervals, totaling 30 measurements per trial. The apparatus used for PD measurement is shown in \autoref{fig:pd_setup}. An LSB200 miniature load cell with a resolution of 1 mN (FUTEK Advanced Sensor Technology, US) is attached on an XYZ linear stage with 1 \micro m accuracy (Newport Corporation, US). The skull is mounted on a vertical column. A hypodermic 304 stainless steel tubing of 0.016'' outer diameter is attached to the load cell to sense contacts with surfaces. Contact positions at the exterior surface of the skull and the wax are recorded when the contact force exceeds 10 mN. The PD result is shown in \autoref{fig:experiment_result}b, where a positive PD value indicates an over-cut and a negative PD value indicates an under-cut. In the teleoperation mode, the maximum over-cut is 2.46 mm and the maximum under-cut is 1.35 mm. In the VF assisted mode, the maximum over-cut is 0.87 mm and the minimum under-cut is 0.96 mm. The mean$\pm$std absolute PD are $0.83\pm0.51$mm for the teleoperation mode and $0.39\pm0.26$mm for the VF assisted mode. The constrained motion enforced by the VF improves the PD result statistically significantly (p-value=7.9E-3). However, there still exist errors due to the registration offset and mechanical deflection in the system. 

The total time to finish the cutting task for the two teleoperation trials are 232s and 237s respectively. For the two VF assisted trials, the total time to finish are 212s and 196s respectively. No statistically significant difference is found between the teleoperated ($234 \pm 3.8$s) and VF assisted ($204 \pm 11$s) modes, with p-value=0.069. Additional trials are needed to further verify the difference of duration between the two modes. 

\section{CONCLUSIONS AND FUTURE WORK}
\label{sec:conclusion}
In this paper, we propose a method for the direct generation of avoidance virtual fixtures from anatomical mesh models. The proposed approach can provide a virtual forbidden area for patient anatomies during robot-assisted surgeries. The method does not assume the convexity/concavity of the anatomy near the tool. The method has been validated in simulation, and the run time experiment is presented. An initial piezoelectric cutting experiment on skull phantoms in a telesurgical environment using the dVRK is reported. Statistically significant improvements have been found in cutting path accuracy and penetration depth control compared to conventional teleoperation. In the future, we plan to demonstrate our method in the presence of anatomical movement and deformation. We also plan to conduct systematic user studies with multiple clinician users.






\bibliography{IEEEfull,ref}{}

\begin{thebibliography}{10}
\providecommand{\url}[1]{#1}
\csname url@samestyle\endcsname
\providecommand{\newblock}{\relax}
\providecommand{\bibinfo}[2]{#2}
\providecommand{\BIBentrySTDinterwordspacing}{\spaceskip=0pt\relax}
\providecommand{\BIBentryALTinterwordstretchfactor}{4}
\providecommand{\BIBentryALTinterwordspacing}{\spaceskip=\fontdimen2\font plus
\BIBentryALTinterwordstretchfactor\fontdimen3\font minus
  \fontdimen4\font\relax}
\providecommand{\BIBforeignlanguage}[2]{{%
\expandafter\ifx\csname l@#1\endcsname\relax
\typeout{** WARNING: IEEEtran.bst: No hyphenation pattern has been}%
\typeout{** loaded for the language `#1'. Using the pattern for}%
\typeout{** the default language instead.}%
\else
\language=\csname l@#1\endcsname
\fi
#2}}
\providecommand{\BIBdecl}{\relax}
\BIBdecl

\bibitem{hunter1976craniosynostosis}
A.~G. Hunter and N.~L. Rudd, ``Craniosynostosis. i. sagittal synostosis; its
  genetics and associated clinical findings in 214 patients who lacked
  involvement of the coronal suture (s),'' \emph{Teratology}, vol.~14, no.~2,
  pp. 185--193, 1976.

\bibitem{tatum2008advances}
S.~A. Tatum and W.~D. Losquadro, ``Advances in craniofacial surgery,''
  \emph{Archives of facial plastic surgery}, vol.~10, no.~6, pp. 376--380,
  2008.

\bibitem{bowyer2013active}
S.~A. Bowyer, B.~L. Davies, and F.~R. y~Baena, ``Active constraints/virtual
  fixtures: A survey,'' \emph{IEEE Transactions on Robotics}, vol.~30, no.~1,
  pp. 138--157, 2013.

\bibitem{ueda2017toward}
H.~Ueda, R.~Suzuki, A.~Nakazawa, Y.~Kurose, M.~M. Marinho, N.~Shono,
  H.~Nakatomi, N.~Saito, E.~Watanabe, A.~Morita \emph{et~al.}, ``Toward
  autonomous collision avoidance for robotic neurosurgery in deep and narrow
  spaces in the brain,'' \emph{Procedia CIRP}, vol.~65, pp. 110--114, 2017.

\bibitem{li2020hybrid}
Z.~Li, M.~Shahbazi, N.~Patel, E.~O’Sullivan, H.~Zhang, K.~Vyas, P.~Chalasani,
  A.~Deguet, P.~L. Gehlbach, I.~Iordachita \emph{et~al.}, ``Hybrid
  robot-assisted frameworks for endomicroscopy scanning in retinal surgeries,''
  \emph{IEEE Transactions on Medical Robotics and Bionics}, vol.~2, no.~2, pp.
  176--187, 2020.

\bibitem{prada2005study}
R.~Prada and S.~Payandeh, ``A study on design and analysis of virtual fixtures
  for cutting in training environments,'' in \emph{First Joint Eurohaptics
  Conference and Symposium on Haptic Interfaces for Virtual Environment and
  Teleoperator Systems. World Haptics Conference}.\hskip 1em plus 0.5em minus
  0.4em\relax IEEE, 2005, pp. 375--380.

\bibitem{park2001virtual}
S.~Park, R.~D. Howe, and D.~F. Torchiana, ``Virtual fixtures for robotic
  cardiac surgery,'' in \emph{International Conference on Medical Image
  Computing and Computer-Assisted Intervention}.\hskip 1em plus 0.5em minus
  0.4em\relax Springer, 2001, pp. 1419--1420.

\bibitem{marinho2019dynamic}
M.~M. Marinho, B.~V. Adorno, K.~Harada, and M.~Mitsuishi, ``Dynamic active
  constraints for surgical robots using vector-field inequalities,'' \emph{IEEE
  Transactions on Robotics}, vol.~35, no.~5, pp. 1166--1185, 2019.

\bibitem{ren2008dynamic}
J.~Ren, R.~V. Patel, K.~A. McIsaac, G.~Guiraudon, and T.~M. Peters, ``Dynamic
  3-d virtual fixtures for minimally invasive beating heart procedures,''
  \emph{IEEE transactions on medical imaging}, vol.~27, no.~8, pp. 1061--1070,
  2008.

\bibitem{li2007spatial}
M.~Li, M.~Ishii, and R.~H. Taylor, ``Spatial motion constraints using virtual
  fixtures generated by anatomy,'' \emph{IEEE Transactions on Robotics},
  vol.~23, no.~1, pp. 4--19, 2007.

\bibitem{santosh2019medical}
K.~Santosh, S.~Antani, D.~S. Guru, and N.~Dey, \emph{Medical Imaging:
  Artificial Intelligence, Image Recognition, and Machine Learning
  Techniques}.\hskip 1em plus 0.5em minus 0.4em\relax CRC Press, 2019.

\bibitem{furuhata2014interactive}
T.~Furuhata, I.~Song, H.~Zhang, Y.~Rabin, and K.~Shimada, ``Interactive
  prostate shape reconstruction from 3d trus images,'' \emph{Journal of
  computational design and engineering}, vol.~1, no.~4, pp. 272--288, 2014.

\bibitem{funda1996constrained}
J.~Funda, R.~H. Taylor, B.~Eldridge, S.~Gomory, and K.~G. Gruben, ``Constrained
  cartesian motion control for teleoperated surgical robots,'' \emph{IEEE
  Transactions on Robotics and Automation}, vol.~12, no.~3, pp. 453--465, 1996.

\bibitem{williams1997augmented}
J.~Williams, R.~Taylor, and L.~Wolff, ``Augmented kd techniques for accelerated
  registration and distance measurement of surfaces,'' \emph{Computer Aided
  Surgery: Computer-Integrated Surgery of the Head and Spine}, pp. 1--21, 1997.

\bibitem{jones19953d}
M.~W. Jones, ``3d distance from a point to a triangle,'' \emph{Department of
  Computer Science, University of Wales Swansea Technical Report CSR-5}, 1995.

\bibitem{kazanzides2014open}
P.~Kazanzides, Z.~Chen, A.~Deguet, G.~S. Fischer, R.~H. Taylor, and S.~P.
  DiMaio, ``An open-source research kit for the da vinci{\textregistered}
  surgical system,'' in \emph{2014 IEEE international conference on robotics
  and automation (ICRA)}.\hskip 1em plus 0.5em minus 0.4em\relax IEEE, 2014,
  pp. 6434--6439.

\bibitem{gordon2018ultrasonic}
A.~Gordon, T.~Looi, J.~Drake, and C.~R. Forrest, ``An ultrasonic bone cutting
  tool for the da vinci research kit,'' in \emph{2018 IEEE International
  Conference on Robotics and Automation (ICRA)}.\hskip 1em plus 0.5em minus
  0.4em\relax IEEE, 2018, pp. 6645--6650.

\bibitem{motherway2009mechanical}
J.~A. Motherway, P.~Verschueren, G.~Van~der Perre, J.~Vander~Sloten, and M.~D.
  Gilchrist, ``The mechanical properties of cranial bone: the effect of loading
  rate and cranial sampling position,'' \emph{Journal of biomechanics},
  vol.~42, no.~13, pp. 2129--2135, 2009.

\bibitem{balasubramanian2011robust}
S.~Balasubramanian, A.~Melendez-Calderon, and E.~Burdet, ``A robust and
  sensitive metric for quantifying movement smoothness,'' \emph{IEEE
  transactions on biomedical engineering}, vol.~59, no.~8, pp. 2126--2136,
  2011.

\end{thebibliography}
\bibliographystyle{IEEEtran}
\end{document}